# Integrating Generative Adversarial Networks and Convolutional Neural Networks for Enhanced Traffic Accidents Detection and Analysis

Zhenghao Xi , *Member, IEEE*, Xiang Liu , *Member, IEEE*, Yaqi Liu, Yitong Cai, and Yangyu Zheng

*Abstract*—Accident detection using Closed Circuit Television (CCTV) footage is one of the most imperative features for enhancing transport safety and efficient traffic control. To this end, this research addresses the issues of supervised monitoring and data deficiency in accident detection systems by adapting excellent deep learning technologies. The motivation arises from rising statistics in the number of car accidents worldwide; this calls for innovation and the establishment of a smart, efficient and automated way of identifying accidents and calling for help to save lives. Addressing the problem of the scarcity of data, the presented framework joins Generative Adversarial Networks (GANs) for synthesizing data and Convolutional Neural Networks (CNN) for model training. Video frames for accidents and non-accidents are collected from YouTube videos, and we perform resizing, image enhancement and image normalisation pixel range adjustments. Three models are used: CNN, Fine-tuned Convolutional Neural Network (FTCNN) and Vision Transformer (VIT) worked best for detecting accidents from CCTV, obtaining an accuracy rate of 94% and 95%, while the CNN model obtained 88%. Such results show that the proposed framework suits traffic safety applications due to its high real-time accident detection capabilities and broad-scale applicability. This work lays the foundation for intelligent surveillance systems in the future for real-time traffic monitoring, smart city framework, and integration of intelligent surveillance systems into emergency management systems.

## I. Introduction

ACCIDENT identification based on CCTV footage is popular to improve the safety of heavy traffic on the road, and detecting accidents automatically is critical to modern safety efforts [1], [2]. The current use of enhanced deep learning methods, especially CNNs and GANs, has enhanced the effectiveness of these detection structures [3], [4]. CNNs effectively identify spatial features in two consecutive frames to detect vehicles and unusual motions such as collisions. Research implementing machine learning models for accident detection has improved substantially; however, most current solutions heavily rely on supervised learning approaches that need extensive labeled training datasets. Accidents occur rarely, which hinders the collection of well-balanced training data. The traditional CNNs experience challenges extracting essential features from complicated accident scenes, reducing their detection performance. Research studies have implemented anomaly detection methods, yet their implementation results in high false alarm rates and limited ability to detect various traffic patterns [5].

Emerging statistics about road traffic accidents confirm the need for advanced smart surveillance systems to improve transportation security and speed up crisis responses [6]. Traditional CCTV monitoring depends on personnel manually monitoring images, yet this approach demands extensive work from staff and contains error possibilities. Deep learning-based computer vision techniques have proven today to be a revolutionary solution to overcome this challenge. The study contributes to the special issue by developing a deep learning model based on Generative Adversarial Networks and Convolutional Neural Networks for traffic accident detection improvement. The synthetic data creation through GANs and the accurate analysis through CNNs and ViTs solves the problem of limited training data available for accident detection. The findings from this research merge machine learning video analytics with intelligent traffic monitoring to advance future-generation surveillance capabilities that protect urban road safety. This proposed framework helps achieve smart urban development by implementing automatic accident monitoring with rapid emergency response systems that form essential parts of contemporary smart transportation systems.

GANs enhance the functioning of CNNs because the generated synthetic data can help expand training datasets and manage issues, such as data paucity and skewed classes [7]. By generating realistic examples from relatively low occurrence rate event distributions, GANs help improve the model's detection of anomalous occurrences in traffic surveillance videos [8]. A semi-supervised approach that implements GANs to predict the next frames in the traffic camera for anomaly detection was presented. This enables the system to predict and later detect accident-prone areas easily [9]. Combined CNN and GAN approaches constitute an efficient means to enable automatic accident identification in real-time analysis with fast reaction capabilities [10]. The strength of

Received 7 February 2025; revised 27 March 2025; accepted 18 May 2025. This work was supported by the National Natural Science Foundation of China under Grant 12104289. The Associate Editor for this article was R. Jhaveri. *(Corresponding author: Zhenghao Xi.)*

The authors are with the School of Electronic and Electrical Engineering, Shanghai University of Engineering Science, Shanghai 201620, China (e-mail: zhenghaoxi@hotmail.com; xliu@sues.edu.cn; ulrich@163.com; 15641257571@163.com; 1256677975@qq.com).

Digital Object Identifier 10.1109/TITS.2025.3573172

both these kinds of systems is that they allow the consumption of large volumes of video data, accurate detection of accidents and appropriate notification of emergency services to minimize response time and likely save lives. When urbanization progresses, and traffic congestion grows, implementing such intelligent surveillance systems is critical for road safety and unperturbed traffic organization [11].

The increasing trend in the number of road accidents at the global level has, therefore, created a very grim reason concerning the establishment of intelligent and autonomous systems that may improve the traffic flow and prevent the causation of more unnecessary casualties. Supervised monitoring of CCTV footage is extremely tiring and risky due to the chances of getting imprecise results owing to exhaustion, which is attributed to the volume of data set in high-density urban areas [12]. However, the problem remains that despite obtaining all of these technological infrastructures which are designed to enhance surveillance, there are no automatic detection systems and thus, even in case of an accident, they cannot promptly rush to the scene and take the accident victims to hospitals and this ends up causing avoidable causalities [13]. Applying deep learning models like CNN and GANs, the automated accident detection system can revolutionise traffic control through efficient and faster accident identification and alerting the appropriate agencies, minimizing confusion and traffic jams resulting from delayed responses.

Deep learning-based accident detection research primarily uses CNNs under supervised learning paradigms to analyse data requiring significant labelled datasets. The scarcity of real-world accident data and unbalanced data collections make it difficult to obtain effective performance from current detection models. Integrating CNNs and GANs solves two essential problems affecting accident detection by finding scarce notable accidents within heavy traffic flow and creating more diverse data for improved overall performance [14]. GANs create artificial accident scenes alongside CNNs, which implement feature identification functions for processed video data. Integrating these techniques provides strong performance when detecting accidents against normal traffic patterns in densely populated urban areas becomes complex. Developing intelligent surveillance systems for future smart cities represents a necessary technological requirement because road accidents cause thousands of worldwide deaths [15]. The proposed research implements AI-based accident detection through a hybrid deep learning system incorporating GANs for data augmentation in conjunction with FTCNN and ViTs to boost feature extraction and classification precision. This method surpasses standard CNN models because it combines ViTs to analyze broad contextual information, which improves anomaly identification capabilities. Researchers have made two significant contributions to the field through this study by linking data augmentation with accident detection and developing a methodology that applies to other anomaly detection tasks, including security monitoring and industrial safety. This research pushes deep learning technology in intelligent transportation systems forward by solving major detection issues, which provides the foundation for the advanced smart surveillance systems of the future.

- **Novel Framework:** Presents a novel framework for automated accident detection, which enhances both the accomplishment and solidity of the technique by combining GANs and CNN. It addresses the problem of the scarcity of data; the presented concept joins GANs for synthesizing data and CNNs for model training. Video frames for accidents and non-accidents are collected from YouTube videos and divided into Training, Validation, and Testing with resizing, image enhancement and image normalization pixel range adjustments.
- **Promising Performance:** FTCNN and vision transformer worked best for detecting accidents from CCTV, obtaining an accuracy rate of 94% and 95%, while the CNN model obtained 88%. Such results show that the proposed framework is suitable for traffic safety applications due to its high real-time accident detection capabilities and broad-scale applicability. This work sows the seed for intelligent surveillance systems in the future for real-time traffic monitoring, smart city framework, and integration of intelligent surveillance systems into emergency management systems.

This paper is divided into five sections to cover all the research aspects of automated accident detection from CCTV footage using GAN and CNN models. Section I focuses on the significance of accident detection in automatically improving road safety, explains the issues of existing systems, and presents the study's goals. Section II summarises previously published articles on the approaches to accident detection, points out the existing methods' weaknesses, and posits the need for more developed models such as GANs and CNNs. Section III provides the approach of using GAN in data augmentation and using the CNN for feature extraction and the classification of accidents, about the architecture and the training of the entire model. The Experimental and Analysis section IV explains the datasets and metrics, and the comparisons and analysis of results prove how effective and robust the model is. Lastly, the conclusion briefly V summarizes the findings, considers implications thereof for practice, acknowledges limitations of the work, and recommends avenues for improvement of the system in future studies.

## II. RELATED WORK

Mashhadi et al. [16] designed a system to facilitate the automated outdoor roadside safety evaluation of rural roads employing GAN-CNN models. The imbalance and scarcity of data required for roadside safety analysis are effectively mitigated using GAN and CNN model fine-tuning data augmentation. The findings prove that integrating a GAN with a CNN yields a more accurate and efficient model than conventional approaches to enhance road safety in rural regions. Sujakumari and Dassan [17] developed a real-time traffic forecasting system in which GAN and Hadoop Distributed File System (HDFS) are suggested to be incorporated. The system is used to determine the appearance of traffic by using CCTV images and videos. In this case, the approach is to feed the system with synthetic traffic data generated by a GAN and use the data to train a prediction model. HDFS is utilized to store and manage all the data produced by the cameras' customers. The findings indicate that the suggested system attains improved assessments of traffic flow estimates.

Nguyen et al. [18] provided an approach for identifying anomalous behaviour of moving objects in surveillance videos is applied in traffic environments. Their approach uses GANs to estimate the subsequent frames in the video stream. To differentiate between the predicted and actual ground truth frames, the system can easily turn into an alarm when it detects abnormal events such as sudden movements from vehicles, cross-signal violations by pedestrians, or any objects falling on the roads. In the proposed approach, the temporal behaviour and spatial distribution characteristics of normal traffic are properly represented to identify disturbances in various areas. The experimental results show that these methods yield high accuracy for anomaly detection while maintaining low complexity. Tahir et al. [19] adopted CCTV video analytics to design and develop a real-time traffic monitoring system. This system helps increase road safety/security by presenting the latest traffic incident report. To achieve this, the authors employ an innovative approach: training a Deep Convolutional Neural Network (FTCNN) model that utilizes the synthetic data for the classification of traffic road events and the creation of summary videos for the classified events. Such an approach limits the amount of storage necessary to deal with visual data while keeping track of all the essential specifics concerning traffic mishaps. It reveals the feasibility of the proposed system in recognizing and categorizing different traffic accidents, including traffic crashes and congestion, on a real-time basis.

Authors in [20] presented a novel neural network structure for detecting outlier events in security video surveillance. To this end, the proposed approach employs normalized attention mechanisms to attend to some features more than others and then recalibrate those features depending on the video content. The analysis explains that the proposed NANN model offers better detection in terms of time and accuracy than existing models in a real-world video surveillance system. Authors in [21] suggested a framework for the traffic incident response, which relies on AI to involve multiple modes to facilitate searching and acting on the incidents. The proposal is based on more specific technologies, including NLP and computer vision. For real-time detection of critical traffic elements and anomalies in traffic videos, it leverages YOLOv11s. Its other components include a vision-language model known as Moondream2 and another large language model. Its purpose is to provide a long descriptive write-up of the scene and a short report on the event with possible recommendations. The findings of this work support the prospect of utilizing this framework to increase the efficiency of traffic control while increasing the population's safety by automating the processes of detecting and responding to accidents.

Authors in [22] presented a method for recognising suspicious action based on a deep autoencoder architecture in surveillance videos. From surveillance videos, the proposed approach extracts features using a CNN autoencoder. Here, the characteristics extracted are employed to train an SVM classifier with "suspicious" and "non-suspicious" categories as the keywords to categorize the videos. These results prove that the suggested approach can effectively provide high accuracy of senior suspicious action identification in surveillance videos. Authors in [23] proposed real-time abnormality detection for the movement of Humans and vehicles, which is addressed in this paper. It presents an innovative approach to Video Data Anomaly Detection. The approach is proposed, therefore, based on deep learning techniques that can capture the normal mode of operations of people and vehicles, respectively. The percentage accuracy figures of the two experiments [sic] indicate that the method proposed in this work is well suited for anomaly detection. The previous work on automated accident detection from CCTV is shown in Table I.

TABLE I
SUMMARY OF RELATED WORK

| Ref. | Focus | Results | Limitations |
|---|---|---|---|
| [16] | Automated roadside safety evaluation using GAN-CNN models. | Enhanced road safety analysis with accurate and efficient models. | Limited to rural roads; potential generalizability issues. |
| [17] | Real-time traffic forecasting with GAN and HDFS. | Improved traffic flow estimation. | Relies heavily on synthetic data; scalability challenges. |
| [19] | Real-time monitoring of road traffic using CCTV analytics. | Classify road traffic events and create summary videos effectively. | When optimizing storage space, important data could be deleted, and event observation will remain restricted. |
| [20] | The system detects outliers in surveillance recordings with Neural Attention Networks. | Helps find more targets faster in video surveillance data. | The attention systems use large amounts of processing power. |
| [21] | An AI Traffic Incident Response System. | Helps manage traffic better and keeps people safe. | Using many AI tools creates problems when connecting them. |
| [22] | Suspicious action recognition with CNN autoencoder. | High accuracy in identifying suspicious actions. | Binary classification limits nuanced understanding. |
| [23] | Anomaly detection in human and vehicle movement. | Suitable method for video data anomaly detection with high accuracy. | Lack of detailed comparative analysis with existing methods. |

In previous research, multiple constraints exist in exploring deep learning-based accident detection techniques. Most detection approaches work strictly with supervised learning techniques even though labelled datasets are scarce due to accident events being rare occurrences. Traditional CNN-based models experience difficulty extracting features from challenging traffic conditions, including low-light environments, and produce elevated false detection rates. Current research employing anomaly detection methods shows limited success because such techniques are unreliable in real-world settings with many systematic false alarms. Due to their restricted capacity to handle various accident situations, existing systems lack synthetic data enhancement techniques. The proposed hybrid deep learning framework developed in this study includes GANs for data enhancement alongside ViTs for superior feature extraction to overcome previous obstacles in accuracy, alongside achieving real-time capabilities.

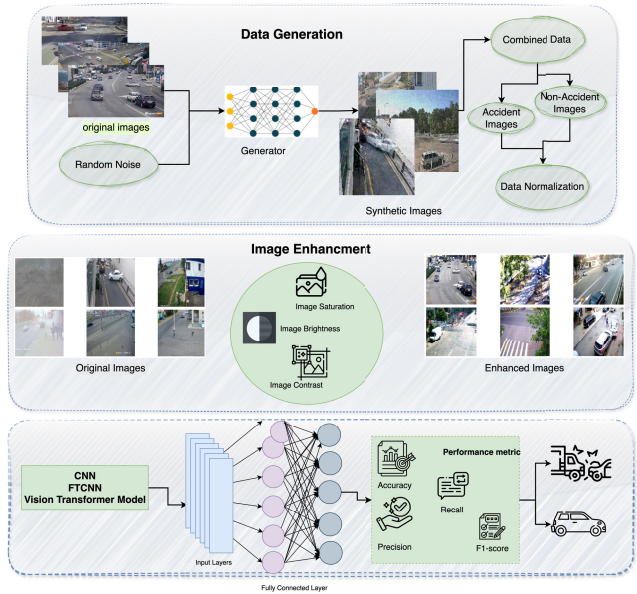

Fig. 1. Proposed framework's workflow.

## III. PROPOSED FRAMEWORK

This paper presents a new and unique framework for accident detection from CCTV images and videos through GANs and CNNs. The proposed framework adopts GANs to generate and improve the data sample's training to overcome the problem of scarce labelled accident data. GANs have two parts: the generator, which is utilised to generate the fake data, and the discriminator, which explains its integrity, whether it is real or fake. The training of these networks in parallel makes GANs produce reasonable accident simulations that enhance the body of data the model has to analyse to generalise across the different forms of accident circumstances that are likely to be observed in reality. Similarly, CNNs are used for feature extraction and classification; therefore, the different accidents are traceable and distinguished from CCTV footage with reasonable accuracy, both accidental as shown in the equation 1 and non-accidental images as shown in equation 2. Combining a GAN to synthesise data and CNNs for enhanced detection of accidents adds a novel perspective that enhances the effectiveness and efficacy of automated reporting. Integrating these techniques offers an optimal framework for accident monitoring in real-time and improving the efficiency and safety of traffic flow. The work sequence of the proposed classifier is described in Fig. 1.

$$x_{\text{accident}} = G(z_{\text{accident}}), \quad z_{\text{accident}} \sim \mathcal{N}(0, 1) \quad (1)$$

where $\mathcal{N}(0, 1)$ is the range between images scaled after generation for both accidental and non-accidental images.

$$x_{\text{non-accident}} = G(z_{\text{non-accident}}), \quad z_{\text{non-accident}} \sim \mathcal{N}(0, 1) \quad (2)$$

The Algorithm 1 of the proposed model of the applied classifier is described below for automated accident detection.

### A. Dataset Preliminaries and Preprocessing

The accident detection from the CCTV footage dataset is made of frames from YouTube videos and characterized by

**Algorithm 1** CNN-Based Accident Classification

1: **Input:** Input image $I$ of dimensions $(H, W, C)$.
2: **Output:** Predicted class $y \in \{0, 1\}$.
3: **Step 1: Convolutional Layer 1** Apply convolution with filter size $(3, 3)$, stride 1, and 32 filters and Apply max pooling having a pooling size of $(2, 2)$:

$$F_1 = \text{ReLU}(I * K_1 + b_1)$$
$$P_1 = \max(F_1)$$

4: **Step 2: Convolutional Layer 2** Apply convolution with filter size $(3, 3)$, stride 1, and 64 filters and max pooling having a pooling size of $(2, 2)$:

$$F_2 = \text{ReLU}(P_1 * K_2 + b_2)$$
$$P_2 = \max(F_2)$$

5: **Step 3: Convolutional Layer 3** Apply convolution with filter size $(3, 3)$, stride 1, and 128 filters and max pooling having a pooling size of $(2, 2)$:

$$F_3 = \text{ReLU}(P_2 * K_3 + b_3)$$
$$P_3 = \max(F_3)$$

6: **Step 4: Flattening Layer** Flatten the 3D feature map $P_3$ into a 1D vector:

$$F = \text{Flatten}(P_3)$$

7: **Step 5: Fully Connected Layers** Apply a dense layer with 256 units and a dense layer as output with 1 unit and sigmoid activation:

$$D_1 = \text{ReLU}(W_1 \cdot F + b_4)$$
$$y = \sigma(W_2 \cdot D_1 + b_5)$$

8: **Step 6: Loss Function**
9: Computing the cross-entropy loss:

$$\mathcal{L} = -\frac{1}{N} \sum_{i=1}^{N} \left[ y_i \log(\hat{y}_i) + (1 - y_i) \log(1 - \hat{y}_i) \right]$$

10: **Step 7: Optimization**
11: Update weights using Adam optimizer:

$$\theta \leftarrow \theta - \eta \nabla_\theta \mathcal{L}$$

12: **Output:** Predicted class $y$.

accident and no-accident frames.[1] The dataset is organized into three folders, partitioned into three types: a training set, a test set and a validation set, each having two subdirectories: the accident and the number of accident frames. To allow models to comprehend sequence and differentiate between accidents and normal traffic flows, consecutive frames of accidents are also incorporated. This structured dataset, which contains images from CCTV frames, provides a framework for generating deep-learning models to detect accidents.

---
[1] https://www.kaggle.com/datasets/ckay16/accident-detection-from-cctv-footage

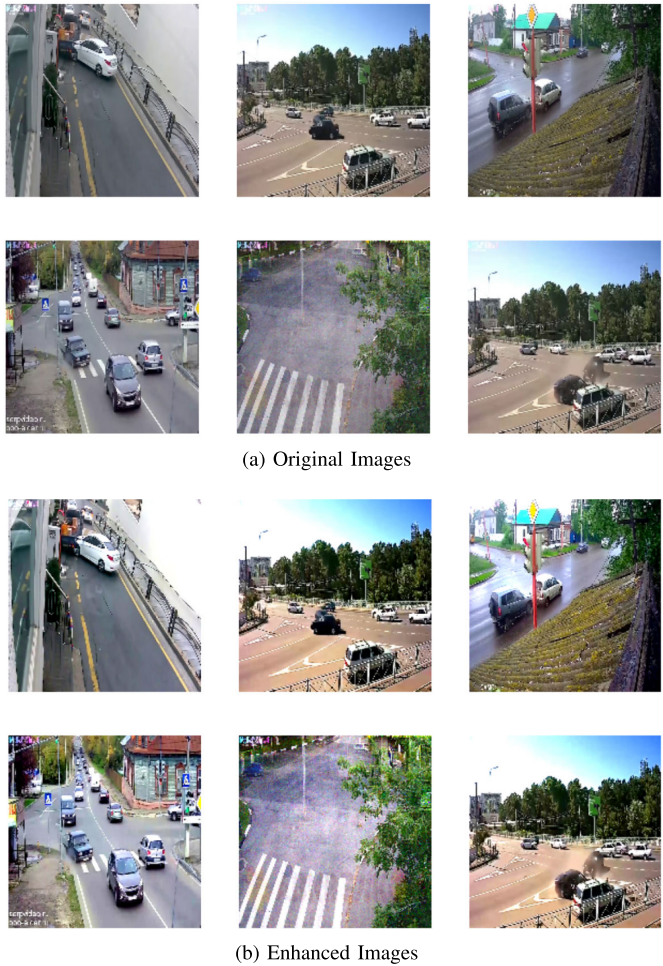

(a) Original Images

(b) Enhanced Images

Fig. 2. Comparison of original and enhanced images.

Several initial actions were taken on the dataset to enhance decisions and improve the model fit. These comprised scaling by resizing the images to have an equal dimension; image enhancement by increasing saturation with the Equation shown in 3 and the brightness as well as the contrast of the images as explained in the equation 4; the enhanced image can be seen in Fig 2 and last was normalization where the pixel values in the images were normalized to enhance the input data seen in the equation 5. Further, to obtain an accurate model evaluation, the data was also divided into training, testing, and validation groups. Together, these steps improve the quality and utility of the dataset for accident identification.

$$\mathcal{D}_{\text{transformed}} = \{(T(x), y) \mid (x, y) \in \mathcal{D}\} \tag{3}$$

$$D_{\text{enhanced}} = \bigcup_{(x,y) \in D} (E(x), y),$$

$$E(x) = \arg\max_{z} \{z \in \mathcal{T}(x) \mid$$

$$z \text{ is optimal under criteria } \Phi(z)\}. \tag{4}$$

$$\mathcal{D}_{\text{normalized}} = \left\{ \left(\frac{x}{255}, y\right) \mid (x, y) \in \mathcal{D} \right\} \tag{5}$$

where $\mathcal{D}_{\text{transformed}}$ represents the enhanced dataset, and Where $\mathcal{D}_{\text{transformed}}$ represents the enhanced dataset, $\mathcal{T}(x)$ is the transformation applied to the image, and $E(x)$ is the enhancement function. and $\mathcal{D}_{\text{normalized}}$ is the dataset after normalization.

### B. Generative Adversarial Network (GAN)

The framework consists of two neural network models, a generator and a discriminator, collaborating to produce realistic synthetic information. We implemented a Deep Convolutional Generative Adversarial Network (DCGAN) to generate synthetic accidents alongside non-accident frames, which enhanced the training dataset accuracy for various scenario detection according to equation 6. A random noise vector z drawn from a normal distribution N(0,1) feeds into the generator network, which uses transposed convolution followed by batch normalization and ReLU activation for high-quality image synthesis. The discriminator network underlines its functionality by applying stride-based convolutional layers and LeakyReLU activations to authenticate or refuse real and synthetic images. Training of the network employs Adam optimization with a learning rate of 0.0002 while beta1 reaches 0.5 value, together with a binary cross-entropy loss function. The DCGAN-based processing technique addresses data limitations while improving model generality, which improves deep learning system accuracy in accident detection as shown by the Algorithm 2.

$$\mathbf{X}_{\text{synthetic}} = G(\mathbf{z}), \quad \mathbf{z} \sim \mathcal{N}(\mu, \sigma^2) \tag{6}$$

where $\mathbf{X}_{\text{synthetic}}$ is the synthetic image generated.

### C. Deep Learning

Deep learning is a branch of machine learning that automatically learns and extracts information from vast volumes of data by using multi-layered artificial neural networks. It allows models to comprehend intricate patterns in text, music, images, and other kinds of data since it is modelled after the structure and operation of the human brain. Deep learning has produced impressive results in applications like image identification, natural language processing, and autonomous systems by utilizing sophisticated architectures like CNNs and RNNs. The deep learning equation 7 is described below.

$$\hat{y} = f(\mathbf{W} \cdot \mathbf{x} + \mathbf{b}) \tag{7}$$

The activation function is represented by f, and w represents the weight matrix.

*1) Convolutional Neural Network:* The configuration for the aforementioned binary classification issues is this CNN model with the subsequent layers. The model uses the Max-Pooling2D layer, which is 2 * 2, after starting with the Conv2D layer, which has 32 3 * 3 filters with ReLU activation. To preserve smaller spatial dimensions while extracting more meaningful features, a similar pattern is continued with additional filter sizes—64 and 128—in the Conv2D layers that follow, which are followed by MaxPooling2D layers. As Table II explains, the feature maps are flattened into a one-dimensional vector and supplied into a Dense or fully connected layer with 256 hidden units with ReLU activation. This CNN model with the following layers is the layout for the binary classification problems as mentioned above. The model starts from the Conv2D layer with 32 filters of 3 * 3 with ReLU activation and then uses the MaxPooling2D layer with a size of 2 * 2. A similar pattern is followed with further filter sizes—64 and 128 — in the subsequent Conv2D layers, followed by MaxPooling2D layers to maintain reduced spatial dimensions but extract more substantial features. Then, the feature maps are flattened into a one-dimensional vector, which is then provided to a Dense or fully connected layer containing 256 hidden units with ReLU activation as described in Table II. However, only one neuron with sigmoid activation, which produces a binary classification, is present in the output layer. The Adam optimizer is displayed in the equation 8, the binary cross-entropy loss function is defined by the equation 9, and accuracy is used as the evaluation metric.

**Algorithm 2** Training Algorithm for DCGAN
**Require:** Training dataset $x$, generator $g$, discriminator $d$, latent space dimension $z$, batch size $b$, number of epochs $e$
1: Initialize $g$ and $d$ with random weights
2: Compile $d$ with Adam optimizer and binary cross-entropy loss
3: freeze $d$ and compile gan model ($g + d$)
4: **for** epoch = 1 to $e$ **do**
5:   **Train discriminator:**
6:   Sample $b$ real images from dataset $x$
7:   Generate $b$ random noise vectors $z$ from $n(0, 1)$
8:   Generate $b$ fake images: $x_{fake} = g(z)$
9:   Assign real labels ($y = 1$) to real images
10:  Assign fake labels ($y = 0$) to fake images
11:  Compute discriminator loss: $l_d = 0.5 \times (d(x) - 1)^2 + 0.5 \times d(g(z))^2$
12:  Update $d$ using back-propagation
13:  **Train generator:**
14:  Generate $b$ new random noise vectors $z'$
15:  Assign all labels as real ($y = 1$)
16:  Compute generator loss: $l_g = (d(g(z')) - 1)^2$
17:  Update $g$ using back-propagation
18:  **if** epoch % 100 == 0 **then**
19:    Print training progress: discriminator loss, generator loss, and accuracy
20:  **end if**
21:  **if** Epoch % save_interval == 0 **then**
22:    Generate new images from $g$ and save results
23:  **end if**
24: **end for**

TABLE II
CNN ARCHITECTURE SUMMARY

| Layer Type | Filters/Units | Activation |
|---|---|---|
| Input Layer | - | - |
| Conv2D | 32 | ReLU |
| MaxPooling2D | - | - |
| Conv2D | 64 | ReLU |
| MaxPooling2D | - | - |
| Conv2D | 128 | ReLU |
| MaxPooling2D | - | - |
| Flatten | - | - |
| Dense | 256 | ReLU |
| Dense (Output Layer) | 1 | Sigmoid |

$$\theta_{t+1} = \theta_t - \alpha \cdot \frac{m_t}{\sqrt{v_t} + \epsilon} \quad (8)$$

where $\theta_t$ is used to represent the model's parameter, and the learning rate is represented by $\alpha$.

$$\mathcal{L} = -\frac{1}{N} \sum_{i=1}^{N} \big[y_i \log(\hat{y}_i) + (1 - y_i) \log(1 - \hat{y}_i)\big] \quad (9)$$

Loss is represented by $\mathcal{L}$, N samples of total data points, and $\hat{y}_i$ represents the predicted output.

*2) Fine-Tuned Convolutional Neural Network (FTCNN):* For complicated tasks like binary classification, the FTCNN architecture is especially well-suited since it allows several convolutional layers to learn and extract characteristics from the data gradually. MaxPooling2D is the second layer, and the first is a Conv2D layer with 32 filters and a size of 3,3, ReLU activation. The following layers deepen, as indicated in Table III, two more Conv2D layers are employed with 64, 128, and 256 filters, respectively. After the convolution and pooling layers, the network goes through a flattened layer, which converts the 3-D feature maps into 1D vectors. To allow the model to train using high-level features, a Dense layer with 512 units and ReLU activation comes after this.

FTCNN architecture is ideal for complex tasks such as binary classification, as several convolutional layers can progressively learn and extract features from the data. The first layer is a Conv2D layer with 32 filters and a size of 3,3, ReLU activation; the second is MaxPooling2D. The subsequent layers expand in depth; another two Conv2D layers are used with 64, 128, and 256 filters, respectively, as shown in Table III. Following the convolution and pooling layers, the network passes through a flattened layer, which reshapes the data from 3-D feature maps to 1-D vectors. This is succeeded by a Dense layer with 512 units and ReLU activation, enabling the model to learn with high-level features. The addition of a Dropout layer with a 50% dropout rate, as outlined in the equation 10, comes before the output layer to reduce overfitting. The output layer has a single neuron that models a binary classification using the sigmoid activation function, as indicated in equation 11 (see also equation 12).

$$\mathcal{L} = -\frac{1}{N} \sum_{i=1}^{N} \big[y_i \log(\hat{y}_i) + (1 - y_i) \log(1 - \hat{y}_i)\big] \quad (10)$$

where the original input feature is represented by $y_i$, and the number of samples is described by N.

$$\theta_{t+1} = \theta_t - \alpha \cdot \frac{m_t}{\sqrt{v_t} + \epsilon} \quad (11)$$

where the learning rate of the classifier is $\alpha$.

$$h = f(\mathbf{W} \cdot (\mathbf{x} \odot \mathbf{m}) + \mathbf{b}) \quad (12)$$

where the weight matrix is w, and the activation function is represented by f.

TABLE III
FTCNN ARCHITECTURE SUMMARY

| Layer Type | Filters/Units | Activation |
|---|---|---|
| Input Layer | - | - |
| Conv2D | 32 | ReLU |
| MaxPooling2D | - | - |
| Conv2D | 64 | ReLU |
| MaxPooling2D | - | - |
| Conv2D | 128 | ReLU |
| MaxPooling2D | - | - |
| Conv2D | 256 | ReLU |
| MaxPooling2D | - | - |
| Flatten | - | - |
| Dense | 512 | ReLU |
| Dropout | - | - |
| Dense (Output Layer) | 1 | Sigmoid |

TABLE IV
ARCHITECTURE OF THE VISION TRANSFORMER (VIT)

| Layer | Output Shape | Param # |
|---|---|---|
| Input | (None, height, width, channels) | 0 |
| Patch Extractor | (None, num_patches, patch_dim) | 0 |
| Reshape | (None, num_patches, patch_dim) | 0 |
| Dense | (None, num_patches, projection_dim) | projection_dim * patch_dim + projection_dim |
| Layer Normalization | (None, num_patches, projection_dim) | projection_dim * 2 |
| MultiHead Attention | (None, num_patches, projection_dim) | projection_dim * num_heads + num_heads$^2$ + projection_dim |
| Add | (None, num_patches, projection_dim) | 0 |
| Layer Normalization | (None, num_patches, projection_dim) | PD * 2 |
| Dense | (None, num_patches, 128) | projection_dim * 128 + 128 |
| Dense | (None, num_patches, projection_dim) | 128 * projection_dim + projection_dim |
| Add | (None, num_patches, projection_dim) | 0 |
| Layer Normalization | (None, num_patches, projection_dim) | PD * 2 |
| Global Average Pooling1D | (None, projection_dim) | 0 |
| Dense | (None, num_classes) | projection_dim * num_classes + num_classes |

*3) Vision Transformer (ViT) Model:* This particular Vision Transformer (ViT) is for the problem of binary classification. The input images are in the shape 224*224*3 and are used as input for the model. The input images are further converted into patches of size 16×16 and are linearly embedded, forming a sequence of patch embeddings. The embeddings are then fed to multiple transformer layers to extract contextual information about the input image, as shown in the equation 16. The model quantifies the projection dimension to be 64 with 4 attention heads and 8 layers of transformers to grasp the spatial and semantic interactions in the picture, as shown in Table IV. A final dense layer with a single neuron and sigmoid activation for binary classification processes the output of the transformer layers. The Adam optimizer, as shown in equation 13, a learning rate of 3e-4, and binary cross-entropy, as indicated in equation 14, are used to train the basic model.

$$\theta_{t+1} = \theta_t - \alpha \cdot \frac{m_t}{\sqrt{v_t} + \epsilon} \quad (13)$$

where the model's parameter is represented by $\theta_t$, and the learning rate by $\theta_t$.

$$\mathcal{L} = -\frac{1}{N}\sum_{i=1}^{N}\left[y_i \log(\hat{y}_i) + (1-y_i)\log(1-\hat{y}_i)\right] \quad (14)$$

where $\mathcal{L}$ and N are the loss and number of samples, respectively, whereas $\hat{y}_i$ is the predicted output.

$$\text{Attention}(Q, K, V) = \text{softmax}\left(\frac{QK^\top}{\sqrt{d_k}}\right)V \quad (15)$$

where $d_k$ represents the key vector's dimensionality (equation 15), the output equation of the VIT classifier is shown in equation 16.

$$\hat{y} = \sigma(\mathbf{W} \cdot h + \mathbf{b}) \quad (16)$$

where $\sigma$ is the activation function, and W is the weight.

## IV. EXPERIMENTAL ANALYSIS AND RESULTS

Then, the experimental and analysis part of the access will examine different Models' performance in the area of automated accident detection from CCTV footage using GANs and CNNs. Generative adversarial networks (GANs) are new-age machine learning architectures for generating new synthetic data that involve two neural network structures: the generator and the discriminator. In this regard, the GANs are applied to create a new realistic accident scenario from the existing dataset to expand the dataset with new labelled examples for improving classifier performance. It assists in countering the issue of scant-labelled accident data by enhancing the training and evaluation of the Models.

Evaluation measures are very important when we want to determine how well our classifier works in making predictions. It aids in quantifying various characteristics of the model, including its accuracy in making the right predictions, its performance on many instances, specifically those of a lesser probability or occurrence, and errors made during the process. Evaluation can further be done based on the models' accuracies, precision, recall, and F1-score, which provides a finer crossover mechanism and the overall comparison to the objectives of a given problem. They also measure the performance by enhancing the model parameters and defining focus areas. These measures show the average percentage of the total number of times a model accurately predicted the results from all the accuracy it achieved, as shown in the equation 17. It is ineffective but straightforward for the datasets with certain classes that dominate them.

$$\text{Accuracy} = \frac{TP}{TP + TN + FP + FN} \quad (17)$$

Precision shows how many of the predictions carried forward as positive actually are positive, shown in the equation 18. It is used in cases where false positives are expensive and time-consuming, such as detecting spam emails.

$$\text{Precision} = \frac{TP}{TP + FP} \quad (18)$$

As can be observed in equation 19, recall is determined by counting the number of real positive cases the model can detect. This is very necessary in applications like medical diagnostic testing, where positives must be detected at any cost.

$$\text{Recall} = \frac{TP}{TP + FN} \quad (19)$$

The precision and recall values, which are constantly traded off one for the other in the equation 20, are used to calculate the F1-score. Because it provides more meaningful results, it is beneficial for examining the model's performance on imbalanced datasets.

$$\text{F1-Score} = \frac{2 \cdot (\text{Precision} \cdot \text{Recall})}{\text{Precision} + \text{Recall}} \quad (20)$$

The ROC Curve provided information about the algorithm's precision and recall performance. The confusion matrix—which included True Positive, True Negative, False Positive, and False Negative rates—was taken into consideration as an additional performance indicator. The amount of actual outcomes of the model compared to the projected outcomes is therefore displayed in a confusion matrix. To identify and fix more specific faults, it displays the total number of true positives and true negatives, false positives, and false negatives for each class. The true positive rate is plotted against the false positive rate using a threshold-dependent graphical form called ROC. The stronger the model, the more up and to the left a curve appears; it will correctly reject many negatives and accurately identify numerous positives. The area under the ROC curve, or AUC, is the primary metric to assess the model's overall performance.

According to the classification reports, three models—CNN, FTCNN, and VIT—performed well on a binary classification job with two classes: "No Accident" and "Accident." Specific metrics from each prediction system, such as precision and recall, the F1-score and overall accuracy, are used to evaluate the model. The neural network architectural design implemented in CNN produced an overall success rate of 88%, which can be seen in Table V. The "No Accident" class demonstrated high precision at 0.82 throughout the classification while achieving 0.96 recall when recognizing actual "No Accident" cases with minor false detection errors. The "Accident" class produced enhanced precision at 0.96 while displaying inferior recall at 0.81 compared to its counterpart. The measurement tools combined at a macro level and weighted value revealed an overall score of 0.89, indicating moderate performance, although the model showed weaker recall ability in detecting "Accident" events.

As seen in Table V, the FTCNN model outperformed the CNN model, achieving 94% accuracy. The "No Accident" category produced high precision (0.92) and recall (0.96) rates, resulting in an F1-score of 0.94. The prediction model identified "Accident" cases with a precise value of 0.96 and a recall realization of 0.92, resulting in an F1-score of 0.94.

TABLE V
CLASSIFICATION REPORT FOR THE CNN, DENSE CNN AND VIT MODEL

| Model | Class | Precision | Recall | F1-Score |
|---|---|---|---|---|
| CNN | No Accident | 0.82 | 0.96 | 0.88 |
| | Accident | 0.96 | 0.81 | 0.88 |
| | Weighted Avg | 0.89 | 0.88 | 0.88 |
| FTCNN | No Accident | 0.92 | 0.96 | 0.94 |
| | Accident | 0.96 | 0.92 | 0.94 |
| | Weighted Avg | 0.94 | 0.94 | 0.94 |
| VIT Model | No Accident | 0.92 | 0.98 | 0.95 |
| | Accident | 0.98 | 0.92 | 0.95 |
| | Weighted Avg | 0.95 | 0.95 | 0.95 |

Several performance indicators show that FTCNN was equally effective in identifying "Accidents" and "No Accidents." The continuous evaluation of macro and weighted averages demonstrated 0.94 as the consistent outcome, which proved the model's effectiveness in managing class imbalance.

The Vision Transformer (ViT) model exhibited superior performance over competing models by reaching overall accuracy at 95%, which is shown in Table V. The "No Accident" class received an F1-score of 0.95 through model performance, demonstrating 0.92 precision and 0.98 recall. According to the analysis, the "Accident" class metrics achieved a precision of 0.98 and a recall of 0.92, resulting in an F1-score of 0.95. With very few wrong results in each category, the model generates outstanding results and performs amazingly well on class predictions. Across the whole dataset, the model performances utilizing weighted averages and macro showed comparable dependability at 0.95.

The VIT model produced the best outcomes among all models by establishing the highest accuracy levels, uniform precision, recall ratings, and F1 scores for both classes. Second, the FTCNN model was used in performance with robust capabilities. The CNN model delivered effective results, although its lower "Accident" recall rate and total accuracy rating imply it would work best for less safety-critical applications. The VIT model exhibits perfect capabilities when precision and recall demands are at their peak.

Loss curves were added to the models to analyze automated accident detection from CCTV footage using GAN and CNN models to reveal important training performance details. During model optimization, a descending loss function across epochs shows that the model moves toward the best possible solution. Both neuron-dense CNN and CNN models achieve efficient learning and good convergence based on their smooth declining loss curve patterns seen in Fig. 3. While the ViT framework began its loss curve trajectory with substantial variations, it currently demonstrates stable declining values representing improved training behaviour. The model has achieved better results after optimization through improved model tuning methods and data preprocessing strategies, in addition to regularization techniques. These curves show additional evidence regarding model success across different training epochs. FTCNN and CNN architecture demonstrate progressive accuracy development in their respective graphs, confirming the models' pattern learning capabilities for CCTV datasets. As depicted in Fig. 3, the ViT model demonstrates consistently recognizable improvements in accuracy throughout each epoch. The comparison shows that the ViT model provides superior data generalization and learning capabilities, which surpass the outcomes of CNN models. The ViT model demonstrates equivalent results compared to other architectural options, indicating its promise in automated accident detection systems.

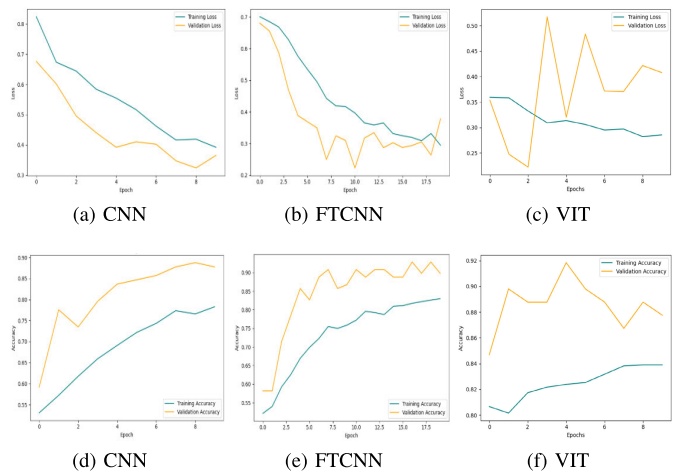

Fig. 3. Accuracy curves of deep learning models.

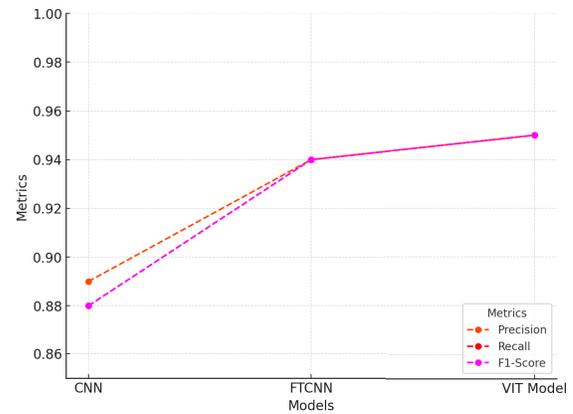

Fig. 5. Performance metrics of applied models.

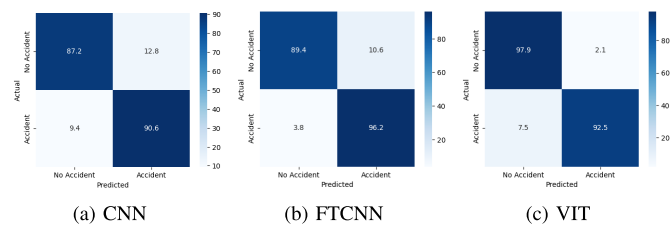

Fig. 4. Confusion matrix of deep learning models.

As seen in Fig. 4, the confusion matrices of both the FTCNN and CNN models show outstanding true positive rates for accidents and non-accident counts, along with low numbers of false positives and negatives. The ViT model achieved enhanced identification accuracy for both traffic accidents and non-accident scenarios. Finally, the model successfully resolves its previous issues with feature extraction and generalization.

This research examined three classification techniques in Automated Accident Detection from CCTV Footage Using GAN and CNN Models. Fig. 5 displays the performance metrics of the suggested models, demonstrating that the Vision Transformer (ViT) model achieved the best results in preventing accidents through CCTV analysis with an accuracy rate of 95%. The ViT model performs better because it successfully models contextual relationships and takes advantage of global dependencies in image data. According to the experimental results, the FTCNN model demonstrated a 94% accuracy rate due to its ability to locate spatial characteristics and features. The baseline CNN model achieved 88% accuracy despite its reliable performance, lagging behind the FTCNN and ViT models. ViT models show superior performance when adequately fine-tuned and provided with adequate data, yet FTCNNs demonstrate persistent effectiveness in image-based accident detection. The study shows that these models demonstrate superior results to traditional CNN architectures, making advanced structures vital for automatic accident detection systems.

## V. CONCLUSION AND FUTURE SCOPE

This paper presented a framework that joins Generative Adversarial Networks (GANs) for synthesizing data and Convolutional Neural Networks (CNN) for model training. The traffic accident detection system combines GANs, CNNs, and ViTs for real-time surveillance. The proposed framework solves the data scarcity issue through GAN-generated synthetic accident simulations, which enhance model stability. ViTs demonstrate 95% accuracy in recognizing abusive accidents in CCTV recordings among standard traffic activities, making the framework highly effective. In the future, we intend to build real-time functionality that qualifies it as an optimal platform for monitoring traffic and emergency response services, thus advancing smart city infrastructure concepts. Moving forward, research must detect a broader range of traffic situations while refining deep learning algorithms for improved precision and connecting this system with IoT-based security protocols to reach commercial deployment potential.

## REFERENCES

[1] A. Rocky, Q. J. Wu, and W. Zhang, "Review of accident detection methods using dashcam videos for autonomous driving vehicles," *IEEE Trans. Intell. Transp. Syst.*, vol. 25, no. 8, pp. 8356–8374, Aug. 2024.

[2] Z. Ning, T. Zhang, X. Li, A. Wu, and G. Shi, "TP-YOLOv8: A lightweight and accurate model for traffic accident recognition," *J. Supercomput.*, vol. 81, no. 4, pp. 1–31, Mar. 2025.

[3] V. A. Adewopo and N. Elsayed, "Smart city transportation: Deep learning ensemble approach for traffic accident detection," *IEEE Access*, vol. 12, pp. 59134–59147, 2024.

[4] Y. Lu, D. Chen, E. Olaniyi, and Y. Huang, "Generative adversarial networks (GANs) for image augmentation in agriculture: A systematic review," *Comput. Electron. Agricult.*, vol. 200, Sep. 2022, Art. no. 107208.

[5] N. Behboudi, S. Moosavi, and R. Ramnath, "Recent advances in traffic accident analysis and prediction: A comprehensive review of machine learning techniques," 2024, *arXiv:2406.13968*.

[6] S. Sai, U. Mittal, and V. Chamola, "DMDAT: Diffusion model-based data augmentation technique for vision-based accident detection in vehicular networks," *IEEE Trans. Veh. Technol.*, vol. 74, no. 2, pp. 2241–2250, Feb. 2025.

[7] S. Xu, M. Marwah, M. Arlitt, and N. Ramakrishnan, "STAN: Synthetic network traffic generation with generative neural models," in *Proc. 2nd Int. Workshop Deployable Mach. Learn. Secur. Defense*, Aug. 2021, pp. 3–29.

[8] K. K. Santhosh, D. P. Dogra, and P. P. Roy, "Anomaly detection in road traffic using visual surveillance: A survey," *ACM Comput. Surv.*, vol. 53, no. 6, pp. 1–26, 2020.

[9] A. Mahbod. (2024). *Advanced Intelligent Monitoring Systems for Traffic Scene Analysis and Anomaly Detection*. [Online]. Available: https://prism.ucalgary.ca

[10] V. S. Rao, R. Balakrishna, Y. A. B. El-Ebiary, P. Thapar, K. A. Saravanan, and S. R. Godla, "AI driven anomaly detection in network traffic using hybrid CNN-GAN," *J. Adv. Inf. Technol.*, vol. 15, no. 7, pp. 886–895, 2024.

[11] K. Ahmad, M. Maabreh, M. Ghaly, K. Khan, J. Qadir, and A. Al-Fuqaha, "Developing future human-centered smart cities: Critical analysis of smart city security, data management, and ethical challenges," *Comput. Sci. Rev.*, vol. 43, Feb. 2022, Art. no. 100452.

[12] M. Ameen and R. Stone, "Advancements in crowd-monitoring system: A comprehensive analysis of systematic approaches and automation algorithms: State-of-the-art," 2023, *arXiv:2308.03907*.

[13] A. Holgersson, "Review of on-scene management of mass-casualty attacks," *J. Human Secur.*, vol. 12, no. 1, pp. 91–111, Aug. 2016.

[14] H. Yijing, W. Wei, Y. He, W. Qihong, and X. Kaiming, "Intelligent algorithms for incident detection and management in smart transportation systems," *Comput. Electr. Eng.*, vol. 110, Sep. 2023, Art. no. 108839.

[15] A. R. Javed et al., "Future smart cities: Requirements, emerging technologies, applications, challenges, and future aspects," *Cities*, vol. 129, Oct. 2022, Art. no. 103794.

[16] A. H. Mashhadi, A. Rashidi, and N. Marković, "A GAN-augmented CNN approach for automated roadside safety assessment of rural roadways," *J. Comput. Civil Eng.*, vol. 38, no. 2, Mar. 2024, Art. no. 04023043.

[17] P. D. Sujakumari and P. Dassan, "Generative adversarial networks (GAN) and HDFS-based realtime traffic forecasting system using CCTV surveillance," *Symmetry*, vol. 15, no. 4, p. 779, Mar. 2023.

[18] K.-T. Nguyen, D.-T. Dinh, M. N. Do, and M.-T. Tran, "Anomaly detection in traffic surveillance videos with GAN-based future frame prediction," in *Proc. Int. Conf. Multimedia Retr.*, Jun. 2020, pp. 457–463.

[19] M. Tahir, Y. Qiao, N. Kanwal, B. Lee, and M. N. Asghar, "Real-time event-driven road traffic monitoring system using CCTV video analytics," *IEEE Access*, vol. 11, pp. 139097–139111, 2023.

[20] V. K. Damera, R. Vatambeti, M. S. Mekala, A. K. Pani, and C. Manjunath, "Normalized attention neural network with adaptive feature recalibration for detecting the unusual activities using video surveillance camera," *Int. J. Saf. & Secur. Eng.*, vol. 13, no. 1, pp. 51–58, Feb. 2023.

[21] A. Ahmed, M. Farhan, H. Eesaar, K. T. Chong, and H. Tayara, "From detection to action: A multimodal AI framework for traffic incident response," *Drones*, vol. 8, no. 12, p. 741, Dec. 2024.

[22] W. Ahmed and M. H. Yousaf, "A deep autoencoder-based approach for suspicious action recognition in surveillance videos," *Arabian J. Sci. Eng.*, vol. 49, no. 3, pp. 3517–3532, Mar. 2024.

[23] I. Pathirannahalage, V. Jayasooriya, J. Samarabandu, and A. Subasinghe, "A comprehensive analysis of real-time video anomaly detection methods for human and vehicular movement," *Multimedia Tools Appl.*, vol. 84, no. 10, pp. 7519–7564, Apr. 2024.

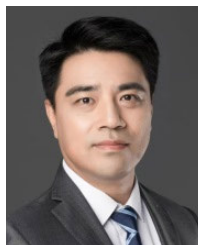

**Zhenghao Xi** (Member, IEEE) received the B.E. degree from South-Central University for Nationalities, Wuhan, China, in 2003, the M.E. degree from the University of Science and Technology Liaoning, Anshan, China, in 2008, and the Ph.D. degree from the University of Science and Technology Beijing, Beijing, China, in 2015. He is currently a Professor with Shanghai University of Engineering Science, Shanghai, China, and the Director of the Department of Automation. His research interests include computer vision, robotics, and intelligent systems.

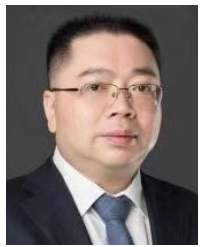

**Xiang Liu** (Member, IEEE) received the B.Sc. degree from Nanjing Normal University, the M.Eng. degree from Jiangsu University, and the Ph.D. degree from Fudan University. He is currently a Professor and the Director of the Computer Department, School of Electronic and Electric Engineering, Shanghai University of Science Engineering, Shanghai, China. His current research interests include computer vision and pattern recognition.

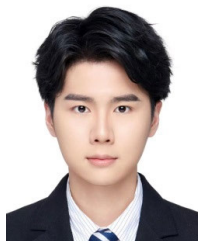

**Yaqi Liu** received the B.E. degree from Shanghai University of Engineering Science, Shanghai, China, in 2024, where he is currently pursuing the M.S. degree. His research interests include computer vision and robotics.

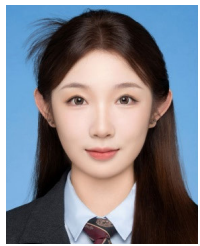

**Yitong Cai** received the B.E. degree from Liaoning Technical University, Huludao, China, in 2024, where she is currently pursuing the M.S. degree. Her research interests include computer vision and robotics.

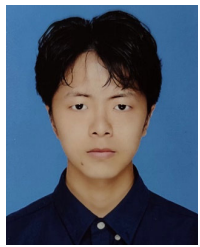

**Yangyu Zheng** received the B.E. degree from Shanghai University of Engineering Science, Shanghai, China, in 2024, where he is currently pursuing the M.S. degree. His research interests include the IoT integration and intelligent transportation systems.